# Resolution enhancement in scanning electron microscopy using deep learning

*Kevin de Haan[1,2,3], Zachary S. Ballard[1,2,3], Yair Rivenson[1,2,3*], Yichen Wu[1,2,3] and Aydogan Ozcan[1,2,3,4*]*

[*] Corresponding authors: rivensonyair@ucla.edu ; ozcan@ucla.edu

[1]Electrical and Computer Engineering Department, University of California, Los Angeles, CA, 90095, USA.

[2]Bioengineering Department, University of California, Los Angeles, CA, 90095, USA.

[3]California NanoSystems Institute (CNSI), University of California, Los Angeles, CA, 90095, USA.

[4]Department of Surgery, David Geffen School of Medicine, University of California, Los Angeles, CA, 90095, USA.

We report resolution enhancement in scanning electron microscopy (SEM) images using a generative adversarial network. We demonstrate the veracity of this deep learning-based super-resolution technique by inferring unresolved features in low-resolution SEM images and comparing them with the accurately co-registered high-resolution SEM images of the same samples. Through spatial frequency analysis, we also report that our method generates images with frequency spectra matching higher resolution SEM images of the same fields-of-view. By using this technique, higher resolution SEM images can be taken faster, while also reducing both electron charging and damage to the samples.

Scanning electron microscopy (SEM) is an important tool for characterization of materials at the nanoscale. By using electrons instead of photons for imaging samples, SEM can achieve sub-nanometer spatial resolution[1], revealing topological and compositional features invisible to traditional light microscopy. Therefore, SEM is frequently employed in a wide range of fields such as material science, biomedicine, chemistry, physics, nanofabrication, and forensics, among others[2–4].

However, when compared to light microscopy, the focused electron beam utilized by SEM is inherently more destructive to samples, especially soft and/or dielectric materials, resulting in electron charge build-up as well as deformation from absorption-based heating[5]. Consequently, these practical barriers prohibit many important samples such as biological specimens, polymers, and hydrogel-structures from being reliably characterized by SEM. There are, however, several approaches to mitigate the destructive effects of the electron beam. For example, it is common practice to coat the samples in e.g., gold, palladium, or iridium prior to imaging[6]. Additionally, shorter dwell times can be used during the electron beam scan to reduce the exposure to the sample. Though helpful, these approaches pose a performance trade-off: to reduce charging effects and sample deformation from heat one must alter the sample from its native state and/or incur increased noise in the acquired image[7].

Although computational approaches for super resolution in electron microscopy have been previously demonstrated[8,9], they require that a portion of the image be taken in high resolution or that the images have similar characteristics and contain sparse unique structures outside of a periodic topology. Other computational enhancements that have been applied to SEM images include denoising as well as deconvolution to reduce the spatial blur in the image caused by the finite beam size[10,11]. Alternative imaging techniques such as ptychography can also be used to increase the resolution in SEM, but these approaches require modification of the imaging set-up[12,13].

Here, we present a deep learning-based approach to improve the resolution of SEM images using a neural network. By training a convolutional neural network (CNN) with a set of co-registered high- and low-resolution SEM images of the same set of samples, we blindly super resolve individual SEM images, reducing sample charging and beam damage without losing image quality or adding extra sample preparation steps. In contrast to previous methods, our approach can be implemented over a wide-range of sample types, and only requires a single SEM image as input. This data-driven approach has the added benefit of reducing the scanning time of the electron beam, and thus increasing the imaging throughput by enabling the use of a lower magnification scan over a larger field-of-view without sacrificing image quality.

Deep neural networks have emerged as an effective method for statistical processing of images and have been shown to improve image quality and achieve super resolution of camera images[14] and across several modalities of optical microscopy[15,16]. Once trained, the network can quickly process input SEM images in a feed-forward and non-iterative manner to blindly infer images with improved quality and resolution, thus making it an attractive and practical tool for rapid SEM image enhancement.

We demonstrated the efficacy of our deep-learning based technique using a gold-on-carbon resolution test specimen [Ted Pella 617-a]. This test specimen has a random assortment of gold nanoparticles of varying sizes ranging from 5 nm to 150 nm immobilized on carbon, and is commonly employed to measure the resolution of SEM systems at different scales using the gaps between various gold nanoparticles.

The image dataset employed to train the CNN was made up of unique high- and low-resolution pairs of the test specimen, each taken from the same region of interest where there is a distribution of nanoparticles. The low-resolution images were taken at a magnification of 10000× (14.2 nm pixel size), while the high resolution images were taken at 20000× magnification (7.1 nm pixel size.) In both cases the image resolution is limited by the number of pixels and therefore the lower magnification images can be modeled as aliased versions of the higher resolution images. A Nova 600 DualBeam-SEM (FEI Company) was used with a 10 kV accelerating voltage, 0.54 nA beam current, and a monopole magnetic immersion lens for high resolution imaging. All images were acquired with 30 μs pixel dwell time.

Once the high- and low-resolution image pairs were taken, they were co-registered before being inputted to the neural network for the training phase. These training images were first roughly matched to each other by cropping the center of each of the low-resolution images and using a Lanczos filter to up-sample the images. After this rough alignment, additional steps were taken to register the images with higher accuracy. First, image rotation and size misalignment were corrected by using the correlation between the two images to define an affine matrix which was then applied to the high resolution images. Next, local registration was performed using a pyramid elastic registration algorithm[17,18]. This algorithm breaks the images into iteratively smaller blocks, registering the local features within the blocks each time, achieving sub-pixel level agreement between the lower and higher resolution SEM images.

40 pairs of accurately registered images (924×780 pixels) were split into 1920 patches (128×128 pixels) which were then used to train the network. The size of the training dataset was further increased by randomly rotating and flipping each image patch. The network model utilized in this work was a Generative Adversarial Network (GAN) which uses a generator network to create the enhanced images, and a discriminator network (D) that helps the generator network (G) to learn how to create realistic high-resolution images[19]. In addition to the standard discriminator loss, an L1 loss term was also added to ensure that the generated images are structurally close to the target, high-resolution images; the anisotropic total variation loss (TV) was also used to increase the sparsity of the output images and reduce noise. Based on this, the overall loss function for the generator network can be written as:

$$l_{generator} = L_1\{G(x), z\} + \alpha \times TV\{G(x)\} + \beta \times [1 - D(G(x))]^2 \quad (1)$$

where $x$ is the low resolution input image to the generator network and $z$ is the matching high resolution ground truth image corresponding to the same field-of-view. α and β are tunable parameters to account for the relative importance of the different loss terms. The $L_1$ loss is the mean pixel difference between the generator's output and the ground truth image, defined as:

$$L_1\{G(x), z\} = \frac{1}{M \times N} \sum_i \sum_j |z_{i,j} - G(x)_{i,j}| \quad (2)$$

where $i$ and $j$ are the pixel indices in an *M×N* pixel image. The anisotropic total variation loss is defined as:

$$TV\{G(x)\} = \sum_i \sum_j \left( |G(x)_{i+1,j} - G(x)_{i,j}| + |G(x)_{i,j+1} - G(x)_{i,j}| \right) \quad (3)$$

The discriminator loss, on the other hand, penalizes the discriminator when it is unable to discriminate between the generated and the ground truth images, and is defined as:

$$l_{discriminator} = D(G(x))^2 + (1 - D(z))^2 \quad (4)$$

The discriminator loss, L1 loss, and the total variation loss make up 84%, 14%, and 2% of the total loss for the generator, respectively. The generator uses an adapted U-net structure[20], while the discriminator uses a modified Visual Geometry Group (VGG) type network structure[21]. Details of these network architectures are shown in Figure 1.

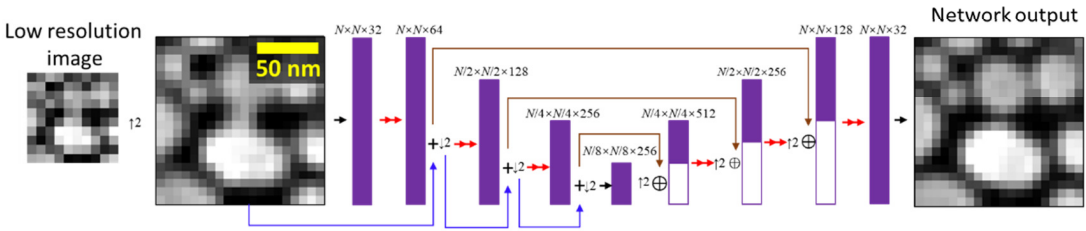

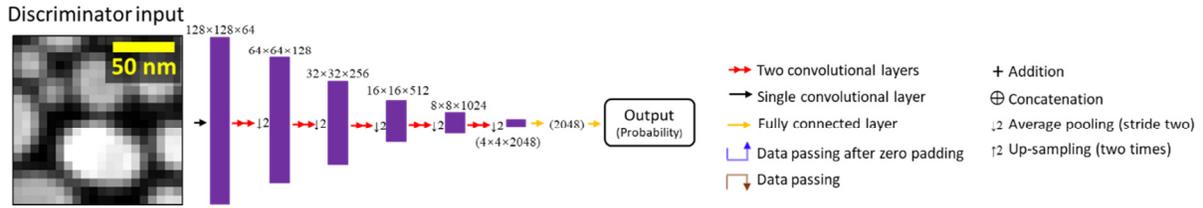

**Figure 1**. Diagram of the network structure. Every convolutional block is made up of two convolutional layers, each followed by a leaky rectified linear unit (ReLU) activation function. The second convolutional layer in each block changes the number of channels. a) The structure of the generator portion of the network. b) The structure of the discriminator portion of the network.

The network was implemented in Python (version 3.6.2) using the TensorFlow library (version 1.8.0). The generator was trained for 48,000 iterations with the discriminator updating every fourth iteration to avoid overfitting. This took the network one hour and twenty minutes to train using a single Nvidia GTX 1080 Ti graphics processing unit (GPU) and an Intel Core i9-7900 processor. The same computer is able to infer 3.66 images per second, for an image size of 780×780 pixels. This inference time is 16 times faster than the low-resolution SEM imaging of the corresponding sample field-of-view; stated differently, real-time visualization of the super-resolved images, immediately after a low resolution image is taken or while a new scan is ongoing, is feasible.

This super resolution technique allows us computationally to enhance the resolution of lower magnification SEM images such that the network's output accurately matches the resolution given by the higher resolution SEM images of the same samples. A demonstration of this can be seen in Figure 2, which reports several blindly tested examples of nanoparticles that are not clearly resolved in the input images, but become distinct after the application of the neural network. Pixel-intensity cross-sections are also reported to illustrate the resolution enhancement more clearly. From these examples we can see that the network is able to reveal spatial details that are not clear in the input (lower magnification) SEM images, matching at its output the corresponding higher magnification SEM images of the same fields-of-view. This is particularly evident in the gaps between the gold nanoparticles shown in Figure 2.

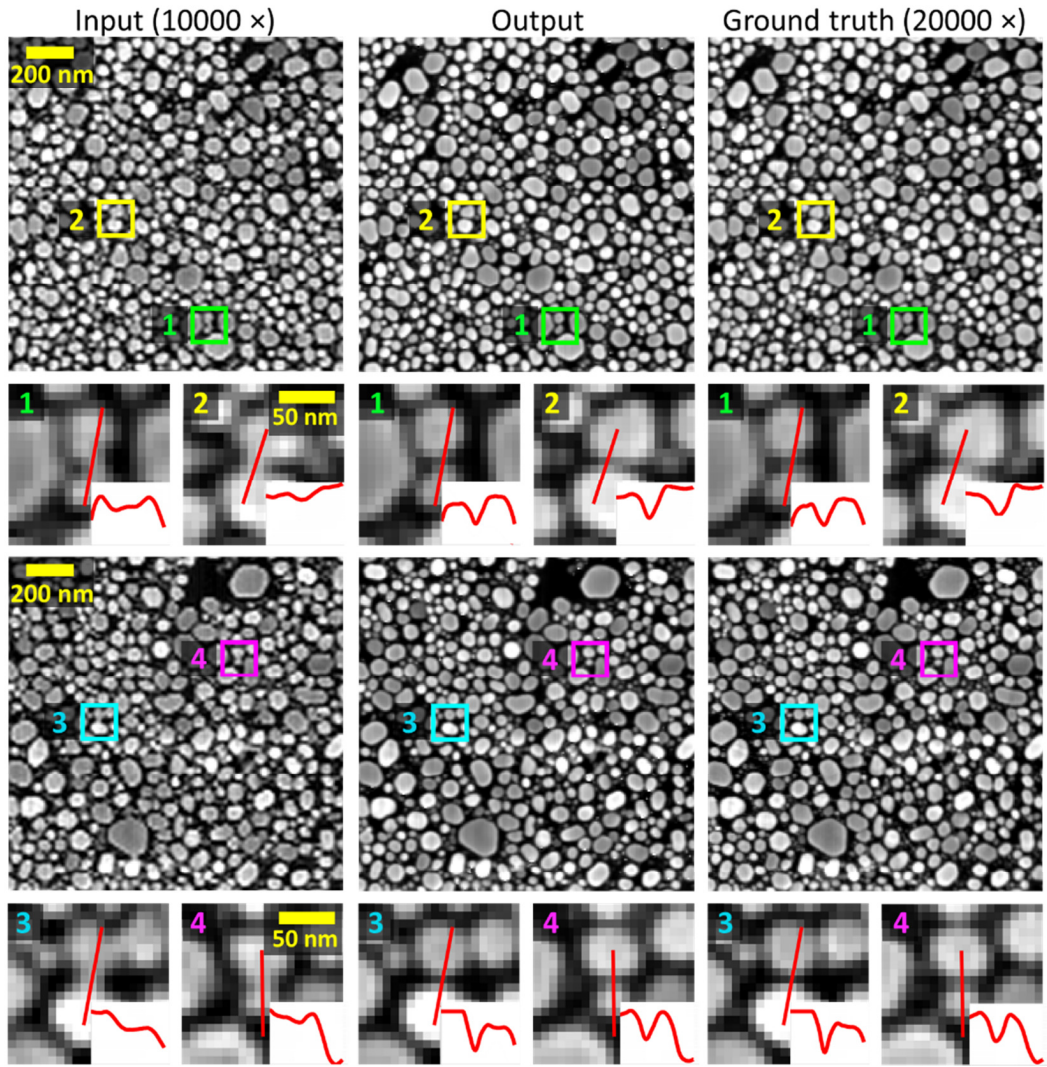

**Figure 2:** Examples of the up-sampled network input images compared to the output and ground truth SEM images. Cross sections of various spatial features with noticeable resolution enhancement are shown.

In fact, Figure 3 provides a statistical analysis of these gaps to quantify the enhancement provided by the trained network; for this analysis, 300 gaps between arbitrary adjacent nanoparticles were randomly selected using the high-resolution SEM images. They were then analyzed to determine whether the neighboring particles are resolvable, as well as to quantify the gap-size in the input image, output image, and target image. The gap width was defined as the distance between the points at which the intensity drops below 80% of highest intensity value of the adjacent particles, and a gap was determined to exist if the lowest intensity point between the particles fell below 60% of the peak value. In the input SEM image (lower magnification), 13.9% of these gaps were not detectible, i.e., could not be resolved (see Fig. 3). However, after super resolving the input SEM images using the trained network, the percentage of undetected gaps dropped to 3.7%. Additionally, the *average difference* between the measured gap sizes in the low- and high-resolution SEM images decreases from 3.8 nm to 2.1 nm after passing through the network.

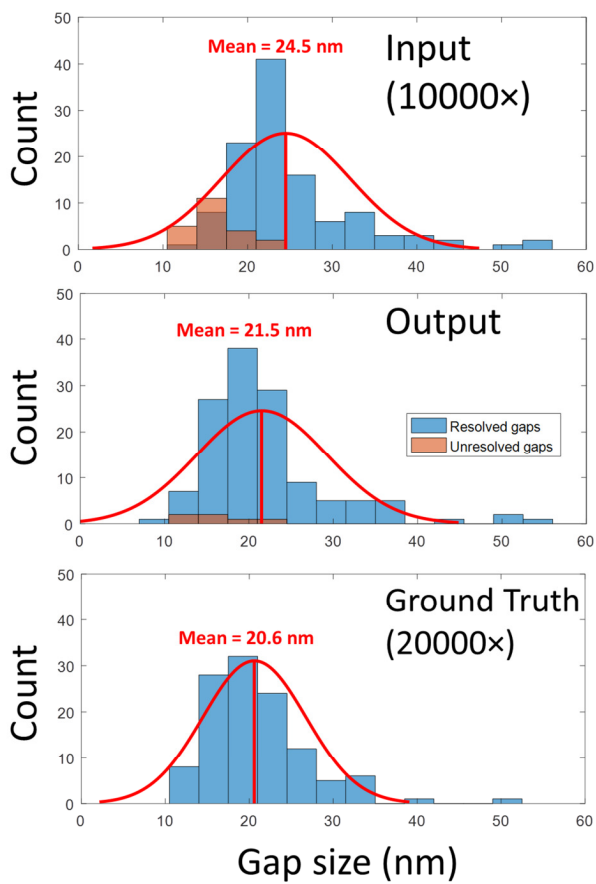

**Figure 3**: Histograms of the gap sizes inferred from the network input and the output images compared to the ground truth image. Total count changes among the histograms due to some of the gaps only being visible in specific images. In the input SEM images, 13.9% of the gaps were not detectible; the percentage of undetected gaps dropped to 3.7% for the output images. A Gaussian distribution, fitted to the gap histograms, with the corresponding mean gap size is also shown for each plot. The number of unresolved gaps in both the input and output images is also shown using a different color; unresolved gaps were not used for mean gap estimation. Pixel size per image is 7.1 nm; the input image is upsampled by a factor of 2.

Another way to illustrate the resolution improvement is reported in the spatial frequency analysis shown in Figure 4. This figure compares the magnitudes of the spatial frequencies for the low- and high-resolution SEM images as well as those of the network output images. From this comparative analysis we can see that the network enhances the high frequency details of the input SEM image such that the spatial frequency distribution of the network output image is consistent with the high-resolution SEM image – including the spatial frequencies that are aliased in the input image due to the large pixel size.

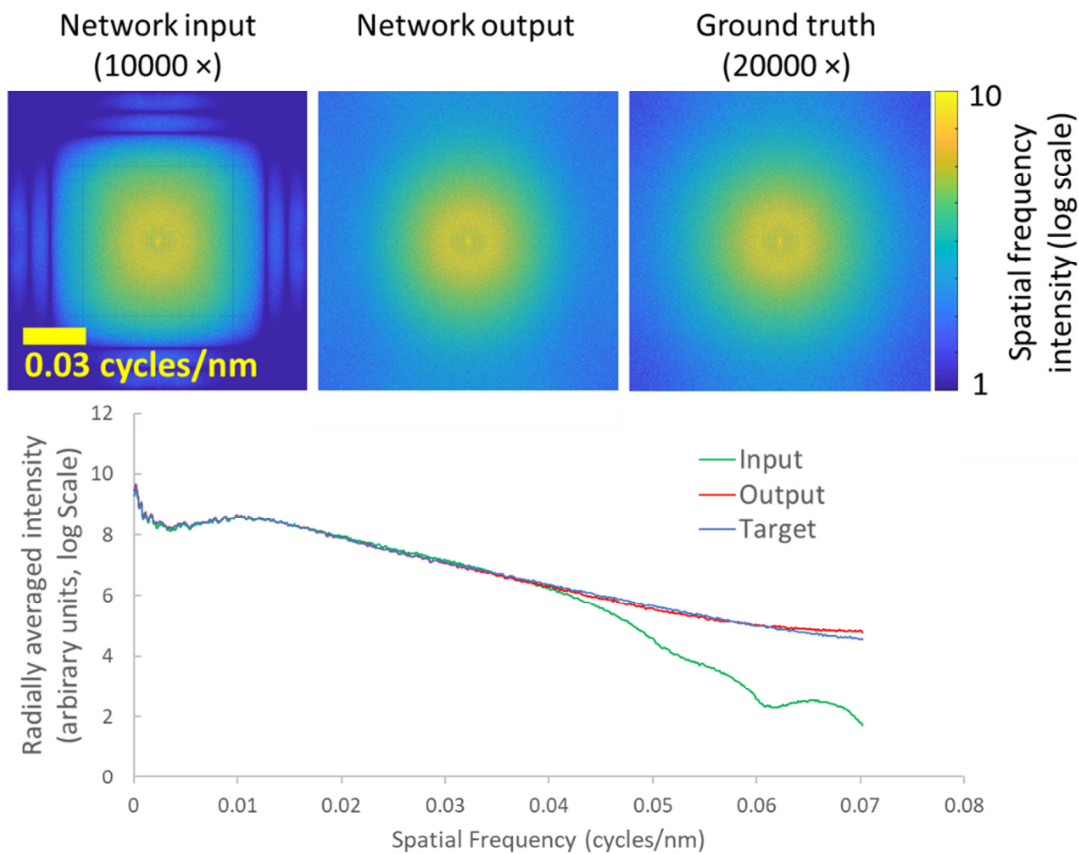

**Figure 4:** Top: spatial frequency distribution of the up-sampled input, output, and ground truth images. Bottom: radially-averaged plot of the above distributions. Analysis was performed on the uncropped versions of the SEM images shown in Figure 2.

Taken together, deep learning-based super resolution is shown to be a powerful and practical tool to computationally improve the resolution in SEM. The 2-fold increase in resolution demonstrated here allows for a *four-fold* reduction of the number of electrons which must interact with the sample to acquire an SEM image, in turn enabling a four-fold increase in the speed of image acquisition. This could significantly benefit the characterization of samples prone to charging or beam-induced damage, by reducing electron exposure without sacrificing image quality. This would allow for higher resolution imaging of a variety of biological materials and nanofabricated samples that previously could not be characterized adequately by SEM.